\documentclass[twocolumn]{apie}

\usepackage{microtype}
\usepackage{graphicx}
\usepackage{booktabs} 
\usepackage{multirow} 
\usepackage{placeins}
\usepackage{listings}
\usepackage{xcolor}

\usepackage{hyperref}

\usepackage{amsmath}
\usepackage{amssymb}
\usepackage{mathtools}
\usepackage{amsthm}

\theoremstyle{plain}
\newtheorem{theorem}{Theorem}[section]

\theoremstyle{definition}
\newtheorem{definition}[theorem]{Definition}

\theoremstyle{remark}

\usepackage[textsize=tiny]{todonotes}

\title{BEAVER: Building Environments with \\Assessable Variation for Evaluating Multi-Objective Reinforcement Learning}

\author[1]{Ruohong Liu}
\author[1]{Jack Umenberger}
\author[2]{Yize Chen}

\affiliation[1]{University of Oxford}
\affiliation[2]{University of Alberta}

\definecolor{verylightgray}{rgb}{0.9,0.9,0.9}
\definecolor{lightblue}{rgb}{0.733,0.875,1.0}
\definecolor{metablue}{rgb}{0, 0.392, 0.898}
\definecolor{forestgreen}{rgb}{0.208,0.667,0.235}
\definecolor{verylightblue}{rgb}{0.839,0.925,1.0}
\definecolor{veryverylightblue}{rgb}{0.92,0.96,1.0}

\abstract{
Recent years have seen significant advancements in designing reinforcement learning (RL)-based agents for building energy management. While individual success is observed in simulated or controlled environments, the scalability of RL approaches in terms of efficiency and generalization across building dynamics and operational scenarios remains an open question. In this work, we formally characterize the generalization space for the cross-environment, multi-objective building energy management task, and formulate the multi-objective contextual RL problem. Such a formulation helps understand the challenges of transferring learned policies across varied operational contexts such as climate and heat convection dynamics under multiple control objectives such as comfort level and energy consumption. We provide a principled way to parameterize such contextual information in realistic building RL environments, and construct a novel benchmark to facilitate the evaluation of generalizable RL algorithms in practical building control tasks. Our results show that existing multi-objective RL methods are capable of achieving reasonable trade-offs between conflicting objectives. However, their performance degrades under certain environment variations, underscoring the importance of incorporating dynamics-dependent contextual information into the policy learning process. BEAVER is open sourced for building RL developments~\footnote{\url{https://github.com/chennnnnyize/BEAVER}}.
}

\date{\today}
\correspondence{Yize Chen \email{yize.chen@ualberta.ca}}

\begin{document}

\maketitle

\section{Introduction}
\label{sec:Introduction}
Consider a building operator who deploys controllers across a portfolio of buildings. Each building may differ in its thermal dynamics due to construction materials, HVAC configurations, or geographic location. Moreover, different groups of occupants may prioritize conflicting objectives—some valuing reduced carbon emissions, while others focus on minimizing electricity bills. In such real-world scenarios, controllers are expected to generalize across unseen building conditions and adapt to diverse user preferences.
\begin{figure*}[ht]
\centerline{\includegraphics[width=0.95\textwidth]{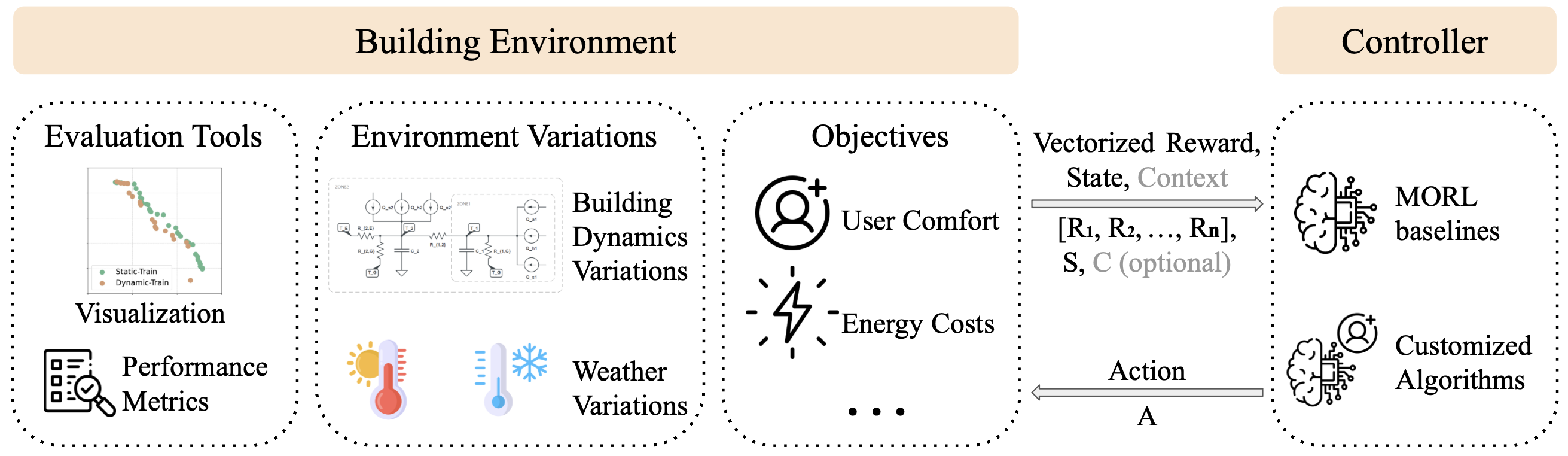}}
\caption{Overview of the BEAVER benchmark framework. The building environment provides multi-objective rewards (e.g., user comfort and energy cost) and supports evaluation under varying building dynamics and climate profiles. BEAVER includes existing MORL baselines and allows integration of customized algorithms. Evaluation tools support both visual and quantitative analysis.}
\label{figure:schematic}
\end{figure*}

To ground learning-based decision-making for building systems, trained agents face similar challenges to those of the building operator: diverse objectives, varying environments~\cite{nweye2023citylearn}. Reinforcement learning (RL) algorithms hold promise in interacting with complex unknown building dynamics and implementing high-dimensional control actions for various HVAC devices. These algorithms have been shown to be effective in optimizing HVAC operations for energy efficiency or occupant comfort~\cite{goldfeder2023lightweight, yu2021review}. However, two key challenges remain underexplored in this context: the ability of RL agents to generalize across diverse building dynamics, and the need to simultaneously optimize multiple, often conflicting objectives. Most existing RL-based controllers require numerous samples, and are trained and evaluated on the same building configuration, which limits their applicability in real-world scenarios where deployment environments may differ~\cite{teoh2025generalization, xu2020one}. Furthermore, real-world building control tasks often require balancing multiple objectives, such as minimizing energy consumption while maximizing indoor comfort~\cite{wang2021multi}. Since each occupant or controller may have distinct preference vectors over these objectives, it is crucial to develop RL agents that are sample-efficient, generalizable, and effective across diverse reward functions and dynamics.

From the RL algorithmic side, existing multi-objective RL (MORL) research largely assumes a single and static environment, while it is unclear how MORL performs across varying underlying dynamics~\cite{liuefficient, yang2019generalized, rame2023rewarded}. Some research discusses transfer learning and generalization in building RL~\cite{fang2023cross, deng2021reinforcement, zhang2020transferable}, for example by pre-training in a source building and adapting a few layers of the policy network to the target building. To advance the practical adoption of RL in building systems, we further identify two concrete research questions: i). What is the space of building generalization and how to appropriately model the space? ii). How do cross-building RL agents behave under varying reward preferences?

In this work, we build upon a realistic, physics-principled RL environment for the building energy management system~\cite{zhang2023bear}, and design the benchmark BEAVER to support multi-objective and generalization evaluation via two important types of context variations for building systems: the variation in \emph{heat convection} (which intrinsically impacts the \emph{underlying state transition}) and \emph{climate zones} (which directly impacts \emph{exogenous states}).  Our proposed method standardizes the evaluation of assessable variations in building RL tasks by automating the process of environment construction conditioned on dynamics and climate variations, reward design incorporating multiple objectives, and RL generalization evaluation. We formalize such a building RL environment as a Multi-Objective Contextual MDP (MOC-MDP), and evaluate generalization using a set of performance metrics. Preliminary evaluations on current MORL algorithms indicate their limited capabilities in addressing diverse variations of building environments. The code is available at \href{https://github.com/chennnnnyize/BEAVER.git}{https://github.com/chennnnnyize/BEAVER.git}.

\section{Multi-Objective Contextual MDP}
\label{sec:Background}

In this Section, we first present the general framework of MORL, followed by a formalization of generalization within the MORL setting. We then describe the evaluation metrics considered in the building MORL tasks. 

\textbf{Multi-objective Markov decision process (MOMDP).} MOMDP is represented by the tuple $\langle\mathcal{S, A, P, R}_{1: n}, \Omega, f, \gamma\rangle$, where $\mathcal{S}, \mathcal{A}, \mathcal{P}(\mathbf{s}_{t+1}|\mathbf{s}_t, \mathbf{a}_t)$, and $\gamma$ represents state space, action space, transition function, and discount factor, respectively. A key distinction between standard MDPs and MOMDPs is that the reward in MOMDPs is an $n$-dimensional vector $\mathbf{r}_t = [\mathcal{R}_1(\mathbf{s}_t, \mathbf{a}_t),\mathcal{R}_2(\mathbf{s}_t, \mathbf{a}_t),\dotsc,\mathcal{R}_n(\mathbf{s}_t, \mathbf{a}_t)]\in \mathbb{R}^n$. Each policy $\pi$ is associated with a vector of expected returns $\boldsymbol{G}^\pi = [G_1^\pi, G_2^\pi, \dots, G_n^\pi]^\top$, where the expected return of the $i^{th}$ objective is given as $G_i^\pi=\mathbb{E}_{\boldsymbol{a}_{t+1} \sim \pi\left(\cdot \mid \boldsymbol{s}_t\right)}\left[ \sum_t \gamma^t\mathcal{R}\left(\boldsymbol{s}_t, \boldsymbol{a}_t\right)_i\right]$ over a given time horizon. We assume all objective returns are observable. The preference space is defined as $\Omega=\{\boldsymbol{\omega} \in \mathbb{R}^n | \sum_{i=1}^{n}\omega_i = 1, \omega_i\geq 0\}$. Given a preference vector $\boldsymbol{\omega}$, the scalarization function $f_{\boldsymbol{\omega}}(\mathbf{r})$ maps a reward vector $\mathbf{r}(\mathbf{s}, \mathbf{a})$ to a scalar utility
\begin{equation}
\label{equ:preference}
    f_{\boldsymbol{\omega}}(\mathbf{r}(\mathbf{s}, \mathbf{a}))=\boldsymbol{\omega}^\top \mathbf{r}(\mathbf{s}, \mathbf{a}).
\end{equation}

Our goal is then to find a multi-objective policy $\pi(\mathbf{a}|\mathbf{s}, \boldsymbol{\omega})$ such that expected scalarized return $\boldsymbol{\omega}^\top \boldsymbol{G}^\pi$ is maximized. 

\textbf{Pareto optimality.} A policy $\pi$ is dominated by policy $\pi^{\prime}$ when there is no objective under which $\pi^{\prime}$ is worse than $\pi$, i.e., $G_i^{\pi}\leq G_i^{\pi^{\prime}}$ for $\forall i \in [1,2,\dotsc,n]$. Pareto optimality is achieved if and only if a policy is not dominated by any other policies. The Pareto set $\Pi_P$ is composed of non-dominated solutions. The corresponding expected return vector $\boldsymbol{G}^\pi$ of policy $\pi\in\Pi_P$ forms Pareto front $P$. Obtaining the Pareto set is often intractable in real-world control problems, and it becomes even more difficult considering the sequential decision-making nature of RL problems. Therefore, MORL aims to obtain a Pareto set $P$ to approximate the optimal Pareto front.

\textbf{Multi-objective contextual Markov decision process (MOC-MDP)} To study generalization in MORL, we adopt the formulation of MOC-MDP as introduced by ~\cite{teoh2025generalization}. MOC-MDP is represented by the tuple $<\mathcal{C, S, A, P, R}_{1: n}, \Omega, f, \gamma, \boldsymbol{M}>$, where $\mathcal{C}$ denotes the context space, and $\boldsymbol{M}$ is a function mapping any $c\in\mathcal{C}$ to a MOC-MDP, i.e. $\boldsymbol{M}(c) = \langle \mathcal{S}, \mathcal{A}, \mathcal{P}^c, \mathcal{R}_{1:n}, \Omega, f, \gamma, \boldsymbol{M} \rangle$. 

In the building RL tasks, $c$ can include climate conditions (temperature, humidity) and building's thermal dynamics (thermal conductivity). This is because even for a building of the same layouts and materials, their state transition and thus underlying MOC-MDP can be affected by different $c$ values, indicating the importance of policy generalizations.


\textbf{Evaluation metrics.} BEAVER supports both quantitative evaluation and visual analysis. For quantitative evaluation, BEAVER key performance metrics, including Hypervolume (HV), Sparsity (SP), and Expected Utility (EU), to assess the generalization capability of each training mode. The detailed definitions of evaluation metrics are provided in Appendix \ref{Appendix: Experiment Setup Details}. Briefly, a higher HV indicates a better approximation of the Pareto front, capturing both convergence and diversity. EU measures the average utility over a predefined distribution of preferences and higher values indicate better overall policy performance. In contrast, a smaller SP reflects a denser set of solutions along the Pareto front.

\section{Benchmark Design: Generalization in Building Control Tasks}
\label{sec:Benchmark}
In this Section, we first describe the building thermal dynamics considered in our setting, followed by an example of multi-objective reward design. We then introduce two specific types of environment variations that are central to the building MORL tasks. Figure \ref{figure:schematic} provides an overview of the BEAVER benchmark, including its support for multi-objective rewards and context variations.

\textbf{Building dynamics.} We adopt the classical Resistance–Capacitance (RC) network to model the thermal dynamics and set up a modular RL environment of a building \cite{zhang2023bear, ma2012predictive}. For a building of $M$ zones, the zonal temperature $T_i$'s dynamics can be described by the following first-order differential equation:
\begin{equation}
\label{equ:RC}
    \begin{aligned}
    C_{i}\frac{dT_i}{dt} 
    =\sum_{j \in \mathcal{N}(i)} \frac{T_{j}-T_{i}}{R_{i, j}}+Q^h_{i}+Q^a_{i}+Q^s_{i};
    \end{aligned}
\end{equation}

where $C_i$ is the thermal capacitance, $R_{i,j}$ is the thermal resistance between Zone $i$ and Zone $j$ and is symmetric, i.e., $R_{i,j}=R_{j,i}$. $\mathcal{N}(i)$ denote Zone $i$'s neighboring zones. If a zone $k$ has an external wall separating the outdoor, there is an additional $R_{k,e}$ term to model the heat transfer between environment temperature $T_e$ and Zone $k$. In addition to the heat transfer between building zones, the three heating variables $Q^{h}_i, Q^{a}_i, Q^{s}_i$ represent the controlled heating/cooling supplied to each zone; the heat gained from indoor occupant activities; and the solar heat gained from windows for Zone $i$ respectively. Nonlinearity is modeled in terms of the solar heat gain considering tilt angle and the shading effects.

\begin{figure*}[h]
\centerline{\includegraphics[width=0.98\textwidth]{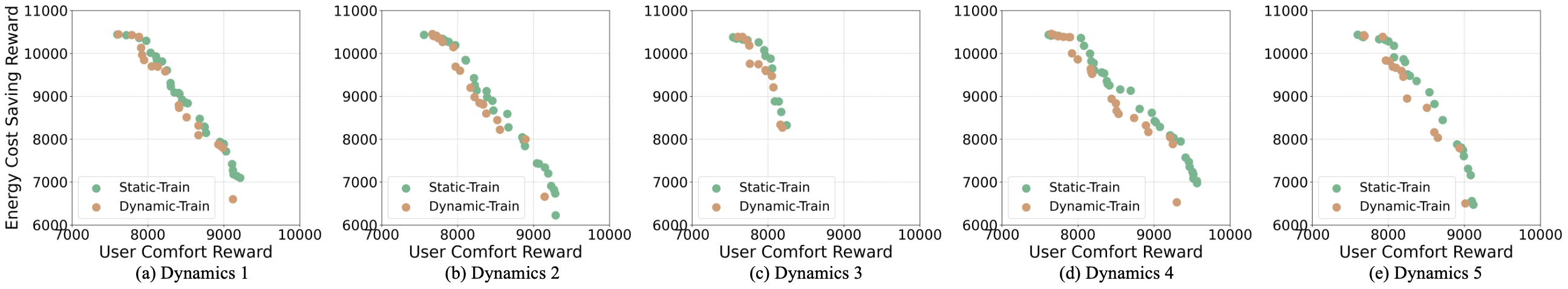}}
\caption{Pareto front comparison on two-objective benchmark in BEAVER.}
\label{figure:pareto}
\end{figure*}

To model building energy management task as a MORL problem, essentially we collect the zonal temperatures as state variables $\mathbf{s}_t$; the controllable heating and cooling inputs as actions $\mathbf{a}_t$; and the set of $\{R,C, Climate\}$ as contextual variables spanning the context space $\mathcal{C}$. 

\textbf{Multi-Objective Reward Design.} As indicated by the example in Introduction, there exist diverse rewards for building RL. Thermal comfort (e.g., temperature deviation from setpoints) and energy consumption are commonly used components in the reward design of many existing benchmarks and control methods for building HVAC systems~\cite{nweye2023citylearn, zhang2023bear, goldfeder2024real}. For the brevity of this paper, we list two examples of customized reward functions at each timestep $t$:

\begin{itemize}
    \item Thermal Comfort: $
\mathcal{R}_{thermal} = M - 0.05 * \sum_i |T_i[t] - T_{i}^s[t]|;$
 where $T_{i}^s$ is user temperature setpoint; 
    \item Energy Cost Savings: $\mathcal{R}_{cost} = M - 0.05 * \sum_i c[t]|P_i[t]|;$ where $c$ is electricity price; $P_i$ is heating/cooling supply power of zone $i$.
\end{itemize}

In BEAVER, we use preference function $f_{\boldsymbol{\omega}}(\mathbf{r})$ defined in \eqref{equ:preference} to map a specific building reward vector $\mathbf{r}(\mathbf{s}, \mathbf{a})$ to a scalar utility given weighted combination using $\boldsymbol{\omega}$. Essentially, $\mathbf{\omega}$ controls the choice of the objectives to be prioritized in the optimization process.

\textbf{Dynamics and Climate Variations.}  To model variations in building's heat dynamics, BEAVER proposes to model the \( U_{\text{wall}} \) parameter for wall's U-factors. In the thermal modeling practice, the U-factor is a measure of a wall's ability to insulate with units of $W/m^2\cdot ^\circ C$. The lower U-factors indicate better insulation and less heat transfer. Indeed, each building has different thermal configurations affected by materials, wall thickness and , leading to varying $C_i$ and $R_i$ parameters in \eqref{equ:RC}. In BEAVER, we follow the reference buildings of EnergyPlus provided by U.S. Department of Energy~\cite{deru2011us}, and sample a group of practical $U_{\text{wall}}$ values, which is further converted to each wall's $R_i, C_i$ values. BEAVER also enables easy user configurations to train and evaluate on varying building dynamics (See Appendix \ref{sec:code}).

Another source of variation in the building MOC-MDP stems from different climate profiles. Such variation is distinct from the RC dynamics variations, in the sense that climate variations do not affect the parameter space of the dynamical systems. Rather, they directly influence the external thermal conditions to which the building is exposed. We parameterize these climate conditions as a set of exogenous weather profile inputs, which can be plugged into the building RL environment. BEAVER provides a set of realistic pre-defined building layouts, weather profiles and $U_{wall}$ values to facilitate comprehensive evaluation of the proposed MOC-MDP framework under various operational constraints and environmental conditions.

Note that in our setting, the context is not observable to the agent. This is because if the context were observable beforehand, it could be concatenated with the state to form an augmented state space, reducing the MOC-MDP to a universal MOMDP~\cite{teoh2025generalization}.

\section{Experiments and Evaluation}
\label{sec:Experiments}

\begin{table*}[htbp]
\caption{Evaluation of HV, EU, and SP for MORL tasks under dynamics variations.}
\label{table:building-2d-dynamics}
\centering
\footnotesize
\setlength{\tabcolsep}{8pt}
\begin{tabular}{lc|ccccc}
\toprule[1.2pt]
Metrics                        & Baselines       & Dynamics 1       & Dynamics 2     & Dynamics 3       & Dynamics 4      & Dynamics 5              \\ \hline
\multirow{2}{*}{HV$(10^{7})$ $\uparrow$}    & Static-Train  & 9.35$\pm$0.04      & 9.38$\pm$0.06     & 8.49$\pm$0.05     & 9.76$\pm$0.05     & 9.27$\pm$0.05 \\
                                 & Dynamic-Train & 9.38$\pm$0.06      & 9.37$\pm$0.07     & 8.59$\pm$0.09     & 9.67$\pm$0.14     & 9.32$\pm$0.09 \\ \hline  
\multirow{2}{*}{EU$(10^{3})$ $\uparrow$}    & Static-Train  & 9.34$\pm$0.01      & 9.33$\pm$0.02     & 9.14$\pm$0.02     & 9.47$\pm$0.02     & 9.32$\pm$0.01 \\
                                 & Dynamic-Train & 9.34$\pm$0.03      & 9.33$\pm$0.04     & 9.15$\pm$0.04     & 9.44$\pm$0.03     & 9.34$\pm$0.03 \\ \hline
\multirow{2}{*}{SP$(10^{5})$$\downarrow$}    & Static-Train  & 0.36$\pm$0.18      & 0.41$\pm$0.14     & 1.27$\pm$1.26     & 0.31$\pm$0.07     & 0.65$\pm$0.42 \\
                                 & Dynamic-Train  & 0.95$\pm$0.46      & 1.04$\pm$0.35     & 1.07$\pm$0.41     & 1.03$\pm$0.56     & 1.13$\pm$0.63 \\ \hline
\bottomrule[1.2pt]
\end{tabular}
\end{table*}

\begin{table*}[!htbp]
\caption{Evaluation of HV, EU, and SP for MORL tasks under climate variations.}
\label{table:building-2d-climate}
\centering
\footnotesize
\setlength{\tabcolsep}{8pt}
\begin{tabular}{c|cccccc}
\toprule[1.2pt]
Metrics       & Mixed Marine       & Cool Marine       & Warm Humid        & Warm Dry          &  Hot Humid        &  Warm Marine$^{*}$              \\ \hline
HV$(10^{7})$ $\uparrow$ & 8.78$\pm$0.11      & 8.74$\pm$0.10     & 8.59$\pm$0.09     & 9.59$\pm$0.05     & 10.00$\pm$0.02    & 10.13$\pm$0.14 \\
EU$(10^3)$  $\uparrow$  & 8.80$\pm$0.06      & 8.78$\pm$0.06     & 8.71$\pm$0.05     & 9.25$\pm$0.04     & 9.58$\pm$0.02     & 9.67$\pm$0.02 \\
SP$(10^{5})$ $\downarrow$ & 0.79$\pm$0.37      & 0.80$\pm$0.30     & 7.24$\pm$0.38     & 0.31$\pm$0.09     & 0.19$\pm$0.08     & 0.19$\pm$0.09 \\ \hline
\bottomrule[1.2pt]
\end{tabular}
\end{table*}

In this Section, we demonstrate how to evaluate generalization performance using our benchmark with state-of-the-art MORL algorithms. We adopt \texttt{C-MORL}~\cite{liuefficient} as our baseline method, which solves a policy optimization problem constrained by a the limits over a group of preference vectors. Studies implemented in \texttt{C-MORL} show strong performance in multi-objective tasks, particularly with a large number of objectives. Our benchmark is also fully compatible with the \texttt{MORL-Generalization} framework~\cite{teoh2025generalization}, which integrates a collection of multi-objective environments (LavaGrid, MuJoCo and etc) with diverse contextual variations and implementations of various MORL algorithms. This allows users to easily evaluate existing methods or their own custom algorithms.

\textbf{Generalization results under dynamics variations.} To evaluate generalization under varying building dynamics, we sample five different sets of $U_{wall}$ parameters, representing buildings constructed with different material properties. These configurations are used to simulate five realistic distinct test environments. We compare performance under two training modes supported by our benchmark: (1) \textit{Static-Train}, where the policy is trained on a single fixed environment; and (2) \textit{Dynamic-Train}, where the policy is trained with a mixture of diverse $U_{wall}$ sampled from a distribution in each training episode.

As shown in Table \ref{table:building-2d-dynamics}, the evaluated baselines demonstrate reasonably stable performance under different dynamics variations. Each result is averaged over 5 runs. For Dynamics 1, 2, 4, and 5, the HV and EU metrics remain relatively high and consistent across both the \textit{Static-Train} and \textit{Dynamic-Train} baselines, indicating good generalization to these environment changes. However, we observe a clear drop in performance under Dynamics 3, where both HV and EU decrease noticeably. Moreover, while \textit{Dynamic-Train} slightly improves HV compared to \textit{Static-Train}, the improvement is marginal, suggesting that the current sampling-based training strategy may not sufficiently enhance robustness to such types of dynamics shifts in buildings. This highlights a key limitation of current MORL approaches, and points to the need for more effective cross-environment RL algorithm design to improve generalization under challenging environment variations.

In addition to quantitative evaluation, BEAVER provides visualization tools such as Pareto front plots (Figure~\ref{figure:pareto}), which enable inspection of algorithmic performance across individual objectives. From the figure, we observe that both \textit{Static-Train} and \textit{Dynamic-Train} perform relatively poorly on the user comfort objective, suggesting suboptimal control solutions of indoor temperature.

\textbf{Generalization results under climate variations.} We evaluate the generalization performance of the MORL agent under climate variations using five different climate profiles. Specifically, the agent is trained in an environment with the Warm Marine climate profile (marked wth $^*$ in Table~\ref{table:building-2d-climate}) and evaluated on the remaining profiles to assess its robustness to unseen climatic conditions. It shows that the agent's performance under different climate variations is relatively unstable. Among the five test profiles, only the Hot Humid climate achieves performance comparable to the training environment, indicating limited generalization across diverse climatic conditions. This indicated that for MOC-MDP, it should provide all necessary information such that a policy optimal in every context can exist and be traced.

\section{Conclusion and Future Directions}
\label{sec:Conclusion}
To bridge the gap between building RL generalization and multi-objective building control, we design BEAVER to automate a practical pipeline for setting up building RL environments and assessing cross-environment RL performance. By accommodating building dynamics and climate zones as context variations, we observe that the performance of baseline algorithms vary significantly across different settings, highlighting the challenge of generalizing MORL policies to unseen environments. To move forward, we are interested in refining BEAVER with more comprehensive building environment variations, such as inclusion of initial value distribution, varying levels of occupancy,  and observable states and etc, which will significantly enhance the practicability of learning-based building controllers. We will also investigate generalizable RL algorithms for buildings.
\bibliography{example_paper}
\bibliographystyle{icml2025}


\onecolumn
\beginappendix
\section{Experiment Setup Details}
\label{Appendix: Experiment Setup Details}
\subsection{MORL Evaluation Metrics}
We evaluate the quality of Pareto front with the following metrics:

\begin{definition}
(Hypervolume). Let $P$ be a Pareto front approximation in an n-dimensional objective space and $\boldsymbol{G}_0$ be the reference point. Then the hypervolume metric $\mathcal{H}(P)$ is calculated as
$
\mathcal{H}(P)=\int_{\mathbb{R}^m} \mathbb{1}_{H(P)}(z) d z,
$
where $H(P)=\{\boldsymbol{z} \in Z|\exists 1 \leq j \leq|P|: \boldsymbol{G}_0 \preceq \boldsymbol{z} \preceq P(j)\}$. $P(j)$ is the $j^{th}$ solution in $P$, $\preceq$ is the relation operator of objective dominance, and $\mathbb{1}_{H(P)}$ is a Dirac delta function that equals $1$ if $z \in H(P)$ and $0$ otherwise. A higher hypervolume is better.
\end{definition}

\begin{definition}
(Expected Utility). Let $P$ be a Pareto front approximation in an $n$-dimensional objective space and $\Pi$ be the policy set. The Expected Utility metric is $\mathcal{U}(P)$
$:
\mathcal{U}(P)=\mathbb{E}_{\boldsymbol{\omega} \sim \Omega}\left[\boldsymbol{\omega}^\top\boldsymbol{G}^{\pi^{SMP}_{\boldsymbol{\omega}}}\right].
$
A higher expected utility is better.
\end{definition} 

\begin{definition}
(Sparsity). Let $P$ be a Pareto front approximation in an $n$-dimensional objective space. Then the Sparsity metric $\mathcal{S}(P)$ is
\begin{equation}
\label{eq:sp}
\mathcal{S}(P)=\frac{1}{|P|-1} \sum_{i=1}^n \sum_{k=1}^{|P|-1}\left(\tilde{G}_i(k)-\tilde{G}_i(k+1)\right)^2,
\end{equation}
where $\tilde{G}_i$ is the sorted list for the $i^{th}$ objective values in $P$, and $\tilde{G}_i(k)$ is the $k^{th}$ value in this sorted list. Lower sparsity is better.
\end{definition}

All three HV, EU, and SP metrics are integrated in MOC-MDP evaluation of BEAVER, which can be called to assess the performance of different policies in the multi-objective setting. 

\section{Modeling of Dynamics Variations}
In BEAVER, the building thermal control environment that controls the temperature of a commercial building with $M$ zones. The states contain the temperature of multiple zones, the heat acquisition from occupant’s activities, the heat gain from the solar irradiance, the ground temperature, the outdoor environment temperature, and the current electricity price $c[t]$. The actions set the controlled heating supply of the zones. The conflicting objectives can be minimizing energy cost, reducing the temperature difference between zonal temperatures and user-set points. We also set up additional reward functions in BEAVER, such as managing the ramping rate of power consumption:

$$
\mathcal{R}_{ramp} = M - |(\sum_i |P_i[t]| - \sum_i |P_i[t-1]|)|.
$$

To model the dynamics variations, we use EnergyPlus as a reference for standard building simulations. To identify the U-wall values, we simulate all reference buildings and find the these distinct $U_{wall}$ values in ``Envelope Summary" table in the main HTML output file generated by EnergyPlus. This table provides a comprehensive overview of the thermal properties of the building envelope, including the U-values for all constructed elements such as walls, roofs, windows, and floors. We make use of EnergyPlus reference buildings to extract $U_{wall}$. They are developed by the U.S. Department of Energy, are a comprehensive set of standardized building models used for energy analysis and simulation. These models represent various commercial building types across different climate zones and are accompanied by extensive documentation and the necessary input files for running simulations in EnergyPlus.

Based on the $U_{\text{wall}}$ extracted from EnergyPlus reference buildings, we define a sampling range for each individual thermal parameter in $U_{\text{wall}}$. During environment construction, we independently sample these U-values within their respective ranges to simulate various building envelopes. While the rest of the building configuration (e.g., layout, HVAC settings) remains consistent with a chosen reference building, this strategy allows us to approximate realistic and cross-building variations. In future work, we plan to expand this sampling approach to additional parameters such as spatial layout and occupancy.


\section{MORL Algorithms}
We choose to implement C-MORL algorithm due to its strong performance and scalability to larger number of objectives~\cite{liuefficient}. In the C-MORL implementation, during the policy initialization stage, we employ parallel training of $M$ policies for pre-known preference vector $\mathbf{\omega}$. We maintain a policy buffer, meaning that in addition to the final policy obtained in the Pareto initialization intermediate policies are also saved in the buffer. We conduct policy selection in both after the Pareto initialization stage and during the Pareto extension stage. Then in the Pareto extension stage, we achieve the goal of filling the Pareto front from selected solutions $\mathcal{X}_{extension}$ toward different directions by solving constrained optimization on the selected policies. Specifically, we consider solving the following problem in which return constraints are controlled by a hyperparameter $\beta \in (0, 1)$:

\begin{equation}
\label{eq:CMOMDP-FEASIBLE}
\pi_{r+1,i}=\arg \max _{\pi \in \Pi_\theta} \{G_{l}^{\pi}: G_{i}^{\pi} \geq \beta G_{i}^{\pi_{r}}, i=1, \ldots, n, i\neq l\},
\end{equation}

We note that most of MORL algorithms to date focus on static environment, where training and evaluation are implemented in a single environment with fixed state transitions or data distributions. While research in multi-task RL and representation learning in RL have proposed to use contextual MDP (CMDP) to model policy transfer, multi-task learning, and environment generalization, previous works focus on single reward preference~\cite{sodhani2021multi, teh2017distral}. This work intends to identify the notable gap in the literature regarding the simultaneous consideration of multi-objectivity and generalization across contexts (i.e. environments or tasks) in building RL tasks.

\section{Discussion of Limitations}
\textbf{Restrictions on State and Action Space.} In our current design of BEAVER, we train and evaluate RL policies with the same dimension of state and action spaces. This design may affect the generalization capabilities of the trained policies, as real-world buildings can have different number of zones and often involve varying but finite dimensions, which can create a mismatch between learned and actual environments. Future work could explore adaptive mechanisms to handle such variability, such as dimensionality reduction techniques and state- or action-level feature embeddings~\cite{bitzer2010using, chandak2019learning}. 

\textbf{Construction of Building Dynamical Models.} Discovering and building up thermal dynamics buildings for building systems is nontrivial, and previous works have focused on model-based approaches to set up high-fidelity model~\cite{crawley2001energyplus, wetter2014modelica}. In the current design of BEAVER, we mainly treat climate and $R,C$ values as two major sources of variations in building models, representing exogenous and dynamical factors respectively. And each building is unique, due to its physical layout, equipment configurations, and location. It is interesting to further customize high fidelity simulation to a specific target building considering detailed building structure, materials, and geographical factors.

\section{BEAVER Additional Implementation Examples}
\label{sec:code}

In this section, we provide a brief introduction to key implementation components of the BEAVER benchmark. The following code snippets illustrate how to instantiate environments, configure dynamic parameters, define climate profiles, and implement vectorized reward functions.


\lstset{
  language=Python,
  basicstyle=\ttfamily\small,
  frame=single,
  numbers=left,
  numberstyle=\tiny\color{gray},
  commentstyle=\color{gray},
  keywordstyle=\color{blue},
  breaklines=true,
  captionpos=b
}

The code below shows how to create a BEAVER environment by generating building parameters and passing them to the simulation environment. Users can flexibly configure simulation conditions, including building type, climate profiles, and locations, through the `ParameterGenerator` interface.
\begin{lstlisting}[caption=Creating a BEAVER environment with building and weather specifications, label={code:env}]
params = ParameterGenerator(Building=args.building, Weather=weather, Location=args.location)
env = BuildingEnv_DR_2d(params)
\end{lstlisting}

This snippet defines the dynamic building components and their corresponding ranges for dynamic parameter sampling (e.g., $U_{Wall}$ values).
\begin{lstlisting}[caption=Sample code for specifying dynamic parameters, label={code:dynamic}]
# Index-to-name mapping for U_Wall elements
self.dyn_ind_to_name = {
    0: 'intwall',
    1: 'floor',
    2: 'outwall',
    3: 'roof',
    4: 'ceiling',
    5: 'groundfloor',
    6: 'window'
        }
        
bounds_range = {
    'intwall': (0.774, 6.299),
    'floor': (0.386, 3.145),
    'outwall': (0.269, 2.191),
    'roof': (0.160, 1.304),
    'ceiling': (0.386, 3.145),
    'groundfloor': (0.386, 3.145),
    'window': (1.950, 3.622)
        }
\end{lstlisting}

BEAVER includes a set of built-in climate files, and users can specify desired weather profiles to evaluate the generalization performance of their learned policies under diverse climatic conditions.
\begin{lstlisting}[caption=Sample code for specifying weather profile, label={code:climate}]
weather_list = ['Mixed_Marine', 'Cool_Marine', 'Warm_Humid', 'Warm_Dry', 'Hot_Humid']
\end{lstlisting}

This example shows how BEAVER environments return multi-objective rewards, including user comfort and saving energy cost, which are computed from indoor temperature difference and energy cost. Users can define custom multi-objective reward functions or select from the predefined options provided by BEAVER.
\begin{lstlisting}[caption=Sample code for defining vectorized reward, label={code:reward}]
def default_reward_function(self, state, action, error, state_new, step_idx):
    reward_comfort = (20*self.roomnum-LA.norm(error, 1))/20
    reward_price = self.roomnum-self.Price_factor*self.ElectricityPrice[step_idx]*LA.norm(action, 1)
    return reward_comfort, reward_price
\end{lstlisting}

\end{document}